\documentclass[letterpaper, 10 pt, conference]{ieeeconf}  

\IEEEoverridecommandlockouts                              

\usepackage{booktabs}                                   
\usepackage{multirow}                                   
\usepackage{makecell}                                   
\usepackage{tablefootnote}                              
\usepackage[symbol]{footmisc}                           
\usepackage{amsmath,amssymb}                            
\usepackage{xcolor}                                     
\let\labelindent\relax              
\usepackage{enumitem}                                   
\usepackage{subcaption}                                 
\usepackage{stfloats}                                   
\usepackage[misc]{ifsym}     
\usepackage{hyperref}   

\usepackage{color}
\usepackage{pifont}

\usepackage[symbol]{footmisc}

\newcommand{\eat}[1]{}                                  

\newcommand{\mL}        {\mathcal{L}}
\newcommand{\mD}        {\mathcal{D}}
\newcommand{\mM}        {\mathcal{M}}

\newcommand{\mS}     {\mathcal{S}}
\newcommand{\mA}     {\mathcal{A}}
\newcommand{\mP}     {\mathcal{P}}

\newcommand{\bbR}     {\mathbb{R}}

\newcommand{\rMTP}     {\mathrm{MTP}}
\newcommand{\rTR}     {\mathrm{TR}}
\newcommand{\rAA}     {\mathrm{AA}}

\newcommand{\rfull}     {\mathrm{full}}
\newcommand{\rpretrain}     {\mathrm{pretrain}}
\newcommand{\rpolicy}     {\mathrm{policy}}

\newcommand{\rmmin}     {\mathop{{\rm min}}}
\newcommand{\rmmax}     {\mathop{{\rm max}}}

\title{\LARGE \bf
Learning from Imperfect Demonstrations with Self-Supervision for Robotic Manipulation
}

\author{Kun Wu$^{1}$, Ning Liu$^{2}$, Zhen Zhao$^{2}$, Di Qiu$^{3}$, Jinming Li$^{4}$,\\ Zhengping Che$^{2}$, Zhiyuan Xu$^{2}$, and Jian Tang$^{2,\dagger}$
\thanks{$^{1}$Syracuse University, NY, USA
        {\tt\small kwu102@syr.edu}}
\thanks{$^{2}$Beijing Innovation Center of Humanoid Robotics, China
        {\tt\small \{Neil.Liu, Alex.Zhao, Z.Che, Eric.Xu, Jian.Tang\}@x-humanoid.com}}
\thanks{$^{3}$Peking University, China
        {\tt\small qiudi@stu.pku.edu.cn}}
\thanks{$^{4}$Shanghai University, China
        {\tt\small ljm2022@shu.edu.cn}}
\thanks{$^\dagger$Corresponding author: Jian Tang.
}
}

\usepackage{caption}
\usepackage{subcaption}
\usepackage{graphicx}  

\begin{document}

\maketitle
\thispagestyle{empty}
\pagestyle{empty}

\begin{abstract}

Improving data utilization, especially for imperfect data from task failures, is crucial for robotic manipulation due to the challenging, time-consuming, and expensive data collection process in the real world.
Current imitation learning (IL) typically discards imperfect data, focusing solely on successful expert data.
While reinforcement learning (RL) can learn from explorations and failures, the sim2real gap and its reliance on dense reward and online exploration make it difficult to apply effectively in real-world scenarios.
In this work, we aim to conquer the challenge of leveraging imperfect data without the need for reward information to improve the model performance for robotic manipulation in an offline manner.
Specifically, we introduce a Self-Supervised Data Filtering framework (SSDF) that combines expert and imperfect data to compute quality scores for failed trajectory segments.
High-quality segments from the failed data are used to expand the training dataset.
Then, the enhanced dataset can be used with any downstream policy learning method for robotic manipulation tasks.
Extensive experiments on the ManiSkill2 benchmark built on the high-fidelity Sapien simulator and real-world robotic manipulation tasks using the Franka robot arm demonstrated that 
the SSDF can accurately expand the training dataset with high-quality imperfect data and improve the success rates for all robotic manipulation tasks.

\end{abstract}

\section{Introduction}

Unlike natural language processing (NLP) and computer vision (CV) tasks, where large volumes of text and image data are easily obtained from the web~\cite{achiam2023gpt,kirillov2023segany}, data for robotic manipulation is scarce and precious due to the high-quality data collection process, especially in real-world applications, is extremely challenging, time-consuming, and costly~\cite{brohan2022rt}.
This makes improving data utilization, particularly incorporating the imperfect data from failed tasks into the model learning process, a critical issue.

As the two dominant learning paradigms, reinforcement learning (RL) and imitation learning (IL) techniques have been extensively researched and applied in the context of robotic manipulation tasks~\cite{brohan2022rt,gu2017deep,zhang2018deep,johns2021coarse,thumm2022provably,chi2023diffusionpolicy} in recent years. However, they still face significant limitations when it comes to utilizing imperfect data.
RL requires an explicit reward function elaborately pre-defined to guide the agent's action. 
Nonetheless, many real-world tasks are intrinsically complicated, making the design of an effective reward function challenging~\cite{ng1999policy,kwon2023reward}.
Furthermore, RL relies heavily on simulators to facilitate low-cost exploration and learning in a trial-and-error manner, which is impractical in real-world settings due to high labor costs, safety concerns, and physical constraints.
On the other hand, IL circumvents the need for explicit reward functions and online exploration by training policy on pre-collected datasets.
This makes IL more feasible for real-world applications~\cite{rahmatizadeh2018vision,jang2022bc,zitkovich2023rt}.
However, IL methods, such as behavioral cloning (BC)~\cite{pomerleau1988alvinn}, operate under the assumption that the data used for training is optimal, which is rarely achievable in practice.
Various factors, such as physical fatigue, task complexity, and line-of-sight occlusion, can lead to task failures during data collection. 
For instance, SayCan~\cite{brohan2023can} collected 276k episodes of data and only retained 12k successful episodes after applying stringent filtering criteria, resulting in more than 90\% of the data being wasted.
In this work, we aim to address the challenge:
\emph{When the reward function and online exploration are unavailable, how can we leverage imperfect data from task failures to improve the data utilization and success rate for robotic manipulation?}

An ideal solution would be to infer a reward function from expert datasets and accurately label all imperfect data. 
Some offline reinforcement learning algorithms~\cite{zolna2020offline,chang2021mitigating,yue2023clare,zeng2024demonstrations,cideron2023get} follow this approach and have achieved promising results in state-based environments. 
However, inferring a reward function without online interaction is challenging. Additionally, most current reward learning algorithms rely on adversarial learning, which is sensitive to hyperparameters, unstable during training, and struggles to learn effectively in high-dimensional environments.
Other IL approaches~\cite{wu2019imitation,sasaki2020behavioral,kim2021demodice,xu2022discriminator,yu2023offline,li2024imitation} focus on learning a policy that mimics the expert's behavior, aiming to mitigate the distribution shift problem~\cite{prudencio2023survey} and improve model performance. 
However, in the context of robotic manipulation, where expert data are scarce and data dimensionality is significantly higher, many methods struggle to align accurately with the data distribution.
For instance, training the discriminator in DWBC~\cite{xu2022discriminator} necessitates costly hyperparameter tuning and struggles with high-dimensional data.
Consequently, there is a need for effective offline IL methods capable of leveraging imperfect demonstrations with high-dimensional input.

Our key insight is motivated by the observation:
\emph{Many steps in a failed trajectory are of high quality despite the overall task failure.}
For example, when attempting to open a cabinet, the robotic arm may successfully grasp the handle but fail to open the cabinet. 
In such cases, the failed data leading up to the successful grasp is still valuable.
More importantly, the expert data from task success certainly contains data that succeeded in grasping the handler, so these two pieces of data will be very similar, providing hints as to how to select high-quality imperfect data.

To this end, we propose a novel \textbf{S}elf-\textbf{S}upervised \textbf{D}ata \textbf{F}iltering framework (SSDF) for robotic manipulation, which extends the training dataset using high-quality imperfect data filtered based on the similarity between the failed trajectory segment and the expert dataset.
Specifically, SSDF contains three main steps.
In step 1, we pre-train a multi-modality transformer network using the mixed demonstrations in a self-supervised manner to extract representative features. 
%
Inspired by pertaining process in BERT~\cite{devlin2018bert}, we propose three self-supervised objectives to enhance the pre-trained transformer, including 1) Masked Transition Prediction, 2) Transition Reconstruction, and 3) Action Autoregression.
These approaches enable the transformer to take on different roles like a forward dynamic model, inverse dynamic model, and behavior cloning model and thus enhance its ability to extract effective features.
In step 2, we aim to identify the useful imperfect demonstrations according to the quality scores and use them to augment the original expert dataset. 
Thus, we leverage the pre-trained transformer in step 1 to extract the features and then generate the quality scores by calculating the similarity between the extracted features of the imperfect and expert demonstrations for each state-action pair in the imperfect demonstrations.
%
%
In step 3, we perform weighted behavior cloning for the high-quality imperfect data, and the original expert dataset can be used for any IL method. 
To differentiate the quality of imperfect demonstrations 
and ensure their beneficial contribution,
we propose the quality scores calculated in step 2 as weights for the behavior cloning loss. 
Our contributions are as follows.
\begin{itemize}
    \item We introduce the Self-Supervised Data Filtering framework (SSDF), which can utilize expert and imperfect demonstrations to train robot manipulation policies.
    \item We design three self-supervised objectives for feature extraction and a novel similarity calculation method, which can accurately expand the training dataset with high-quality imperfect data.
    \item Extensive experiments on the ManiSkill2 and real-world tasks demonstrated that SSDF can effectively utilize imperfect demonstrations to improve performance.
\end{itemize}

\section{Related Work}

Imitation Learning has enormous potential to meet critical safety and cost desiderata for real-world applications.
As an important class of methods, Behavior Cloning (BC)~\cite{pomerleau1988alvinn} can be used in the offline setting seamlessly and gain comparable performance compared to GAIL methods~\cite{ho2016generative}.
IBC~\cite{florence2022implicit}, ACT~\cite{zhao2023learning}, and Diffusion Policy~\cite{chi2023diffusionpolicy} further enhance BC using the network with better expressive capability and advanced training objectives.
However, most works are based on the assumption that all demonstrations are optimal, thus dropping their performance with imperfect data.
In addition, when scaling up the size of datasets in the real world, such as OpenX~\cite{o2024open}and RoboMIND~\cite{wu2024robomind}, it becomes harder to guarantee that the datasets are all optimal as they get larger.

Some offline reinforcement learning algorithms~\cite{zolna2020offline,chang2021mitigating,yue2023clare,zeng2024demonstrations,cideron2023get} follow this approach and have achieved promising results in state-based environments. 
However, inferring a reward function without online interaction is challenging. Additionally, most current reward learning algorithms rely on adversarial learning, which is sensitive to hyperparameters, unstable during training, and struggles to learn effectively in high-dimensional environments.

Offline inverse reinforcement learning (Offline IRL) focuses on training a discriminator in an adversarial manner~\cite{Kostrikov2020Imitation,sun2021softdice,swamy2021moments,jarboui2021offline} or learning a reward function~\cite{zolna2020offline,yue2023clare,zeng2024demonstrations,cideron2023get} to match the distribution of the expert demonstrations.
For example, MIIRL~\cite{zeng2022maximum} proposes to train a conservative world model for learning the reward function from the expert data.
Nonetheless, the adversarial optimization process in these methods can lead to training instability and collapse, which is exacerbated in high-dimensional environments.
In this work, we aim to conquer the challenge of leveraging imperfect data in the context of robotic manipulation with high-dimensional inputs.

Many IL methods with imperfect demonstrations~\cite{wu2019imitation,wang2018supervised,brantley2019disagreement,brown2019extrapolating,brown2020better,tangkaratt2020variational,du2023behavior} are also proposed.
Nonetheless, most methods require extra labeling or online interaction to provide more information to learn.
More recently, several works~\cite{kim2021demodice,yu2023offline,xu2022discriminator,li2024imitation} have been proposed to solve the more challenging problem of offline imitation learning with imperfect demonstrations.
ISW-BC~\cite{li2024imitation} proposes an importance-sampling-based method to re-weigh imperfect data and implicitly make the learned policy close to the expert policy. 
Some other imitation learning methods leveraging powerful architectures~\cite{janner2021offline,shafiullah2022behavior}, such as transformers~\cite{vaswani2017attention}, can flexibly integrate diverse data. 
These methods typically use goal-conditioned training~\cite{chen2021decision, cui2022play} or pretraining~\cite{bonatti2023pact,wu2023masked,sun2023smart} to extract more representative features. 
However, they are not specifically designed to handle imperfect data.
In contrast, our method designs a tailored self-supervised framework for computing the quality scores of imperfect data and is applied to vision-based robotic manipulation.

\section{Methodology}

\begin{figure*}[htp]
    \centering
    \includegraphics[width=0.95\textwidth,trim=0 130 0 130,clip]{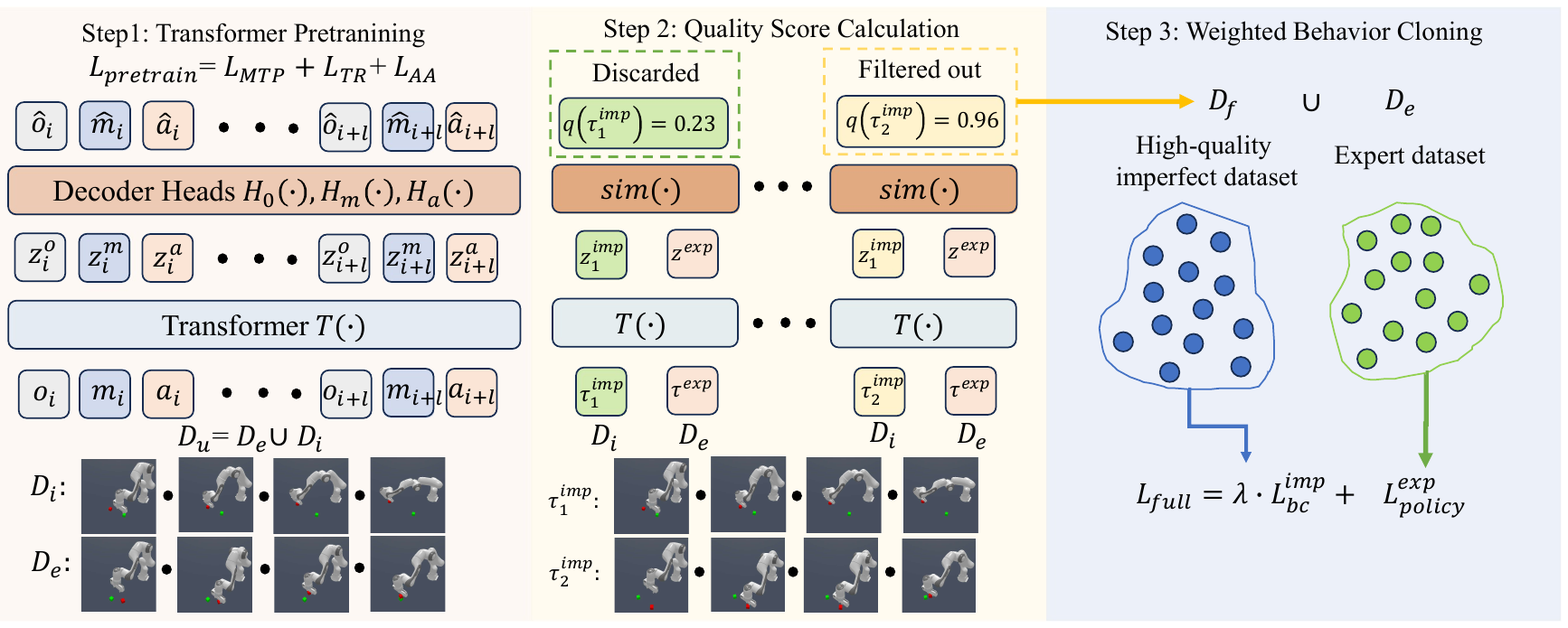}
    \caption{Overview of SSDF. SSDF contains three steps: 1) Transformer Pretraining via Self-Supervised Learning, 2) Calculation of Quality Score by Similarity Metric, and 3) Policy Learning with Weighted Behavior Cloning.}
    \label{fig:overview}
\vspace{-15pt}
\end{figure*}

\subsection{Preliminary on Imitation Learning}

We consider a standard fully observed Markov decision process (MDP)~\cite{sutton1998introduction} to model the environment. 
The MDP $\mM = (\mS, \mA, \mP, r, \rho_0, \gamma)$ contains the state space $\mS$, action space $\mA$, 
reward function $r: \mS \times \mA \times \mS \rightarrow \bbR$, 
state transition function $\mP : \mS \times \mA \times \mS \rightarrow [0,1]$, 
the initial state distribution $\rho_0 (s)$,
and discount factor $\gamma \in (0,1)$.
In the IL setting with imperfect demonstrations, the reward function $r$ is unavailable.
Instead, we have a static expert demonstration dataset $\mD_{e} = \{ (s, a, s') \}$ and a static imperfect demonstration dataset $\mD_{i} = \{ (s, a, s') \}$.
%
%
Our goal is to maximize the success rates for the tasks using both the expert dataset and imperfect dataset $\mD_{u} = \mD_{e} \cup \mD_{i}$ without any reward information and online interaction with the environments.

\subsection{ Self-Supervised Data Filtering Framework}
The overview of the proposed method Self-Supervised Data Filtering framework (SSDF) is depicted in Figure~\ref{fig:overview}.
Specifically, SSDF comprises three main steps.
(1) Transformer Pretraining via Self-Supervised Learning. (2) Calculation of Quality Score by Similarity Metric. (3) Policy Learning with Weighted Behavior Cloning.

\subsection{Transformer Pretraining via Self-Supervised Learning}
\label{sec:pretrain_tf}
\begin{figure}[t]
    \centering
    \includegraphics[width=0.85\linewidth]{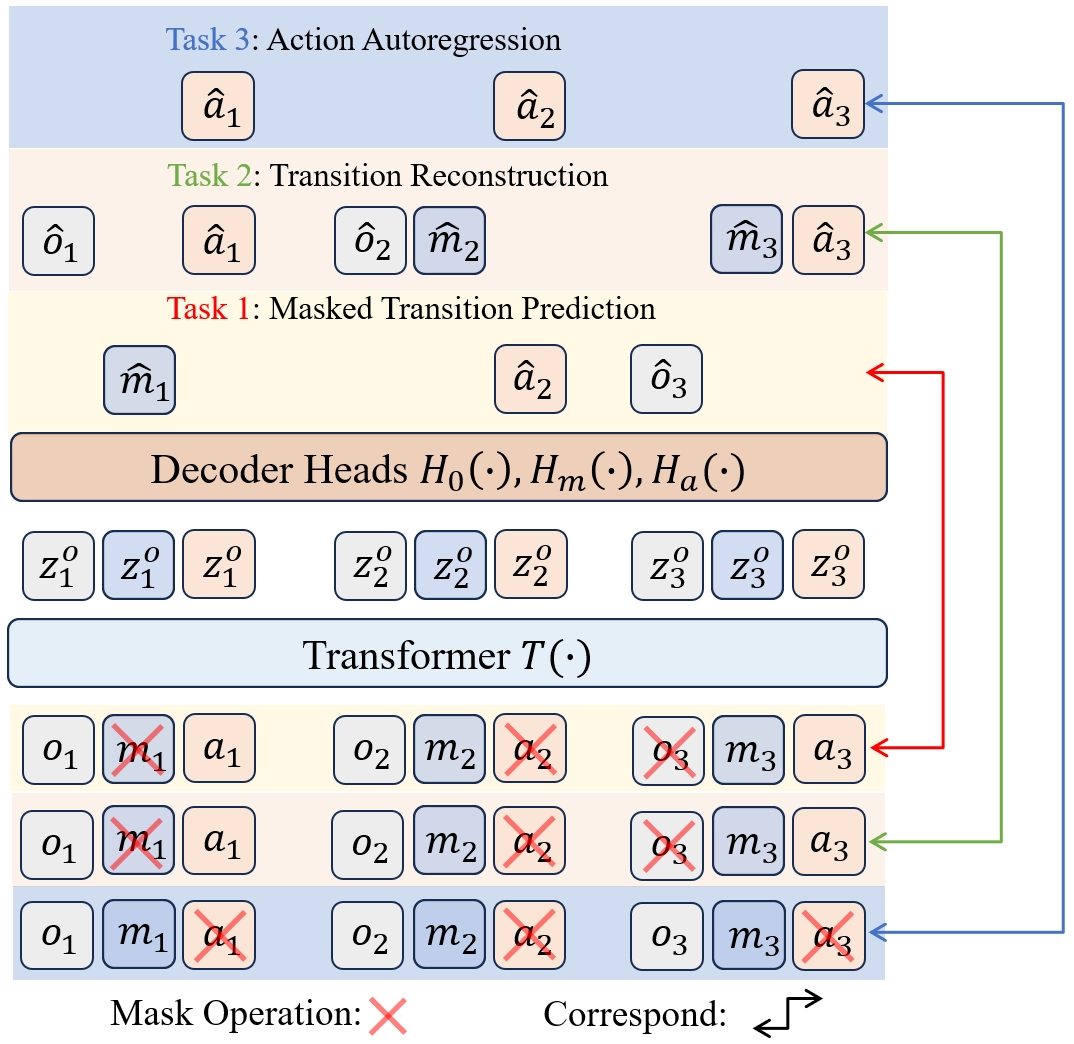}
    \caption{The transformer pre-training process includes three tasks: 1) Masked Transition Prediction (MTP), 2) Transition Reconstruction (TR), and 3) Action Autoregression (AA). Here is an example of the pre-training process with three time-step inputs. Colored lines link the input and output.}
    \label{fig:pretrain}
\vspace{-15pt}
\end{figure}

We leverage the transformer as our base model since it is suitable for modeling long dependencies.
To pre-train the model, we propose three self-supervised approaches to pre-train the transformer using the union dataset $\mD_{u} = \mD_{e} \cup \mD_{i}$.
As shown in Figure~\ref{fig:pretrain}, the three tasks are 
1) Masked Transition Prediction, 
2) Transition Reconstruction, 
and 3) Action Autoregression.
For all tasks, the multi-modality transformer network $T(\cdot)$ takes a trajectory segment $\tau_{in} = {(o_{i}, m_{i}, a_{i}), \cdots, (o_{i+l}, m_{i+l}, a_{i+l})}$ as input and outs the corresponding features $\tau_{out} = {(z^{o}_{i}, z^{m}_{i}, z^{a}_{i}), \cdots, (z^{o}_{i+l}, z^{m}_{i+l}, z^{a}_{i+l})}$ for all input elements, where $i$ is the start index of the time step, $l$ is the segment length, $o_{t}$ is the RGBD images given from the depth cameras, $m_{t}$ is the robot proprioceptive states and necessary information like goal position, $a_{t}$ is the action.
Then we have three separate decoder heads $H_{o}(\cdot), H_{m}(\cdot), H_{a}(\cdot)$ for the above three kinds of modalities, which takes the corresponding features $z^{o}_{i}, z^{m}_{i}, z^{a}_{i}$ as input and outputs the reconstruction or prediction results $\hat{o}_{i}, \hat{m}_{i}, \hat{a}_{i}$.

For Masked Transition Prediction (MTP) tasks, we randomly mask a part of the input trajectory segment with a pre-defined probability (randomly chosen from ${0.4, 0.3, 0.2, 0.1}$ in our implementation), which is denoted as $M(\tau)$.
We denote the optimized parameters in the transformer network $T(\cdot)$ and the decoder heads $H_{o}(\cdot), H_{m}(\cdot), H_{a}(\cdot)$ as $\theta$.
And the goal is given by:
\vspace{-10pt}
\begin{gather}
\label{equ: mtp_loss}
    \mL_{\rMTP} = \rmmax \limits_{\theta} \mathbb{E}_{\tau} \sum_{t=i}^{i+l} \mathbb{I}(e_t) log P_{\theta} (e_t | M(\tau)),
\vspace{-7pt}
\end{gather}
where $e_t \in \{o_{t}, m_{t}, a_{t} \}$ is the input element of the three modalities, $\mathbb{I}(e_t)$ is an indicator function showing the input element is masked (i.e., 1) or not (i.e., 0).
By randomly masking input elements and predicting them, we encourage the transformer network to learn various roles, including forward dynamic model, inverse dynamic model, and data generating policy that boost the ability of feature extraction.

For Transition Reconstruction (TR) tasks, we randomly mask a part of the input trajectory segment and reconstruct the unmasked elements.
The goal is given by:
\vspace{-5pt}
\begin{gather}
\setlength{\abovedisplayskip}{8pt}
\setlength{\belowdisplayskip}{8pt}
\label{equ: tr_loss}
    \mL_{\rTR} = \rmmax \limits_{\theta} \mathbb{E}_{\tau} \sum_{t=i}^{i+l} (1 - \mathbb{I}(e_t)) log P_{\theta} (e_t | M(\tau)).
\vspace{-5pt}
\end{gather}
By reconstructing the unmasked input elements, the transformer network would learn to compress the key information and extract more representative latent features.

To be consistent with the final objective of the robotic manipulation, which is to output action, 
we propose the Action Autoregression (AA) task that forces the transformer network to predict the next action based on the history transitions.
The goal is given by:
\vspace{-7pt}
\begin{gather}
\setlength{\abovedisplayskip}{8pt}
\setlength{\belowdisplayskip}{8pt}
\label{equ: aa_loss}
    \mL_{\rAA} = \rmmax \limits_{\theta} \mathbb{E}_{\tau} \sum_{t=i}^{i+l} log P_{\theta} (a_t | his(a_t)).
\vspace{-10pt}
\end{gather}
where $his(a_t) = (o_{i}, m_{i}, a_{i}), \cdots, (o_{i+t}, m_{i+t} )$ is the history transitions (i.e., we use causal mask in our implementations).

By combining the above three self-supervised tasks, the final pre-training step objective is as follows:
\vspace{-5pt}
\begin{gather}
\label{equ: pretrain_loss}
    \mL_{\rpretrain} = \mL_{\rMTP} + \mL_{\rTR} +  \mL_{\rAA}.
\vspace{-2pt}
\end{gather}

\subsection{Calculation of Quality Score by Similarity Metric}
\label{sec:sim}

Since the expert demonstrations have high quality, we believe that the imperfect demonstrations that is more similar to expert demonstrations is of higher quality.
After pre-training the transformer network in a self-supervised manner, the transformer network now has an enhanced ability to extract expressive and representative features $z^{o}_{i}, z^{m}_{i}, z^{a}_{i}$ for the input elements.
Thus, we choose to use the output features of the expert demonstrations and the imperfect demonstrations to calculate the similarities and final quality scores.
More specifically, for each trajectory segment $\tau^{imp}$ in the imperfect demonstrations, we extract the output features $z^{o,imp}_{i+l}, z^{m,imp}_{i+l}, z^{a,imp}_{i+l}$ at the last time step.
We also extract the output features $z^{o,exp}_{i+l}, z^{m,exp}_{i+l}, z^{a,exp}_{i+l}$ of each expert trajectory segment $\tau^{exp}$ in the same way.
The similarity between the imperfect features and the expert features
are defined as the negative $L_2$ distance:
\begin{gather}
\label{equ: cal_sim}
    sim(\tau^{imp}, \tau^{exp}) = - || z^{o,imp}_{i+l} - z^{o,exp}_{i+l} ||_{2} \notag
    \\
    - || z^{m,imp}_{i+l} - z^{m,exp}_{i+l} ||_{2} - || z^{a,imp}_{i+l} - z^{a,exp}_{i+l} ||_{2}.
\end{gather}
\vspace{-3pt}
The reason why we choose the features of the last time step $z_{i+l}$ of the trajectory segment is that generally the last time step is the most important step for predicting the next action, and the attention mechanism in the transformer has extracted the historical information.
For each imperfect trajectory segment, we need to calculate the similarities with all other expert trajectory segments and the similarity result $w(\tau^{imp})$ is the highest one, which is as follows:
\vspace{-3pt}
\begin{gather}
\label{equ: sim_results}
    w(\tau^{imp}) = \rmmax sim(\tau^{imp}, \tau^{exp}), \ \ \forall \tau^{exp} \in \mD_{e}.
\end{gather}
Once we get similarities of all imperfect segments, we obtain the quality scores by normalizing them:
\vspace{-2pt}
\begin{gather}
\label{equ: qs_results}
    q(\tau^{imp}) = norm(w(\tau^{imp})), \ q(\tau^{imp}) \in [0,1]
\end{gather}

\subsection{Policy Learning with Weighted Behavior Cloning}
\label{sec:wbc}

After calculating the quality scores for all imperfect trajectory segments $\tau^{imp}$, we can rank them 
and filter the high-quality part out to learn the optimal policy.
In order to prevent low-quality imperfect demonstrations from interfering with policy learning, we define a threshold $\beta$ and choose the imperfect segments $\tau^{imp}$
as follows:
\vspace{-3pt}
\begin{gather}
\label{equ: sim_results}
    \mD_{f} = \{ \tau^{imp} | q(\tau^{imp}) > \beta \}
\end{gather}
For the reserved high-quality dataset $\mD_{f}$, we use their quality scores as the weights for behavior cloning.
Combined with the expert demonstrations $\mD_{e}$, the final objective for downstream policy learning is given by:
\vspace{-2pt}
\begin{gather}
\label{equ: sim_results}
    \mL_{\rfull} = \rmmin \limits_{\theta^{'}} \mL_{\rpolicy} - \lambda \mathbb{E}_{\tau \sim {\mD_{f}}} [q(\tau)  log P_{\theta^{'}} (a_{i+l} | \tau)],
\end{gather}
where $\mL_{\rpolicy}$ is the objective for any downstream policy model parameterized by $\theta^{'}$ on the expert dataset $\mD_{e}$, $\lambda$ is a hyperparameter to balance the training of the expert demonstrations and imperfect demonstrations.
We finetune the transformer in our simulation experiments and use ACT~\cite{zhao2023learning} in our real-world experiments.
We can count the distribution of quality scores $q(\tau)$ offline and then adjust the value of $\beta$ accordingly.
We set $l=6, \lambda=0.1, \beta =0.9$ in our implementation.

\section{Experiments}

\begin{figure*}[tp]
\centering
\includegraphics[width=0.95\linewidth, trim=0 180 0 180,clip]{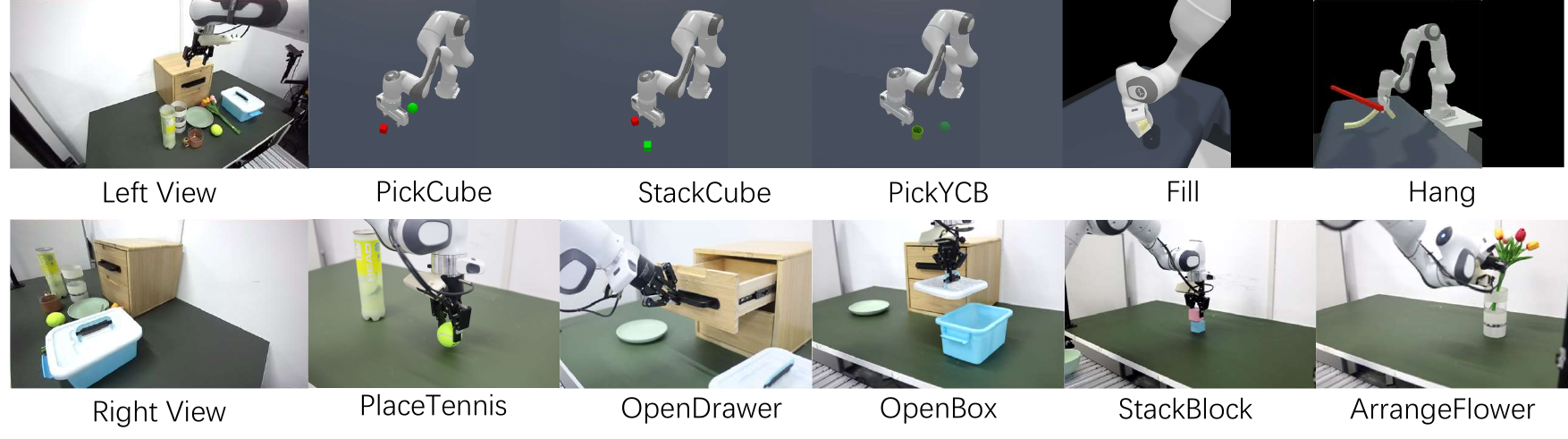}
\caption{Experiment Setup: we conducted experiments on five tasks on the ManiSkill2 benchmark and five tasks using the real-world single Franka robotic arm.
Fill and Hang are soft-body tasks.
The green circles in PickCube and PickYCB represent the goal position.}
\label{fig:task_setting}
\vspace{-8pt}
\end{figure*}

\subsection{Experiment Setup}

\textbf{Simulation Experiments.} Our simulation experiments are based on the ManiSkill2 benchmark~\cite{gu2023maniskill2}, which is built on the high-fidelity Sapien simulator~\cite{xiang2020sapien} and allows agents to be trained using static demonstrations.
The tasks include three rigid object manipulation tasks and two soft-body tasks as shown in Figure~\ref{fig:task_setting}. 
For data collection, we trained different level behavior agents using the online DAPG+PPO~\cite{rajeswaran18dapg,schulman2017proximal} method provided in the ManiSkill2.

\noindent
\textbf{PickCube-v0} is to pick up a red cube and move it to a specified endpoint.
We collected 300 expert trajectories and 900 imperfect trajectories consisting of three groups of 300 trajectories, which are collected by three level agents with success rates of 0.0, 0.46, and 0.91, respectively.

\noindent
\textbf{StackCube-v0} is to pick up a red cube and place it onto the green one.
We collected 300 expert trajectories and 900 imperfect trajectories consisting of three groups of 300 trajectories, which are generated by three level agents with success rates of 0.0, 0.60, and 0.86, respectively.

\noindent
\textbf{PickYCB-v0} is to pick up a YCB object \cite{calli2015ycb} and move it to a specified endpoint.
We selected 10 objects from all objects that are relatively easy to pick up.
We collected 300 expert trajectories and 300 other imperfect trajectories from an imperfect agent with success rates of 0.41.

\noindent
\textbf{Fill-v0} is to fill the target beaker with clay in a bucket.
400 expert trajectories and 400 imperfect trajectories are collected from an imperfect agent with success rates of 0.35.

\noindent
\textbf{Hang-v0} is to hang a noodle on a specified stick.
We collected 400 expert trajectories and 400 imperfect trajectories from an agent with success rates of 0.14.

For the PickYCB, Fill and Hang tasks, we only collected one level of imperfect demonstrations because the behavior policies with high success rates are not available. 
For all tasks, the robot initialization pose, object position, and goal position are randomly generated for each episode.
The robot is a 7-DoF Franka Panda robot with a parallel-jaw gripper.
The input RGBD images $o \in \bbR^{2*128*128*4}$ are from two cameras, including a hand-eye camera and a top fixed camera.
The input robot proprioceptive states $m \in \bbR^{38}$ include joint positions, joint velocities, robot base position, goal position, and end-effector position.
The output action $a \in \bbR^{7}$ is the delta target end-effector pose.
We use the success rate to measure the performance of an algorithm.
Please refer to the Maniskill2~\cite{gu2023maniskill2} for more details.
%

\begin{table}[t]
    \centering
    \caption{Comparisons in terms of success rates of the five tasks on ManiSkill2 benchmark. BC and TF-BC are the baselines that only use expert demonstrations. All other baselines used imperfect demonstrations.}
    \label{tab:maniskill-main-result1}
    \resizebox{\columnwidth}{!}{
        \begin{tabular}{c|ccccc}
        \toprule
        Method & PickCube & StackCube & PickYCB & Fill & Hang \\ \midrule
        BC~\cite{pomerleau1988alvinn} & 82.6 & 80.0 & 43.6 & 24.8 & 14.4 \\
        TF-BC~\cite{vaswani2017attention} & 82.2 & 81.6 & 42.4 & 25.2 & 16.0 \\
        TT~\cite{janner2021offline} & 22.0 & 20.0 & 12.4 & 14.8 & 20.4 \\
        BeT~\cite{shafiullah2022behavior} & 82.6 & 81.2 & 34.6 & 10.6 & 18.6 \\
        \midrule
        PACT~\cite{bonatti2023pact} & 82.0 & 83.2 & 38.4 & 14.6 & 16.0  \\
        MTM~\cite{wu2023masked} & 66.6 & 79.4 & 32.0 & 23.2 & 18.4 \\
        SMART~\cite{sun2023smart} & 65.4 & 81.2 & 25.2 & 8.8 & 16.2  \\
        \midrule
        DT~\cite{chen2021decision} & 30.8 & 78.8 & 36.8 & 4.4 & 6.0 \\
        C-BeT~\cite{cui2022play} & 78.6 & 83.6 & 28.0 & 20.0 & 16.4 \\
        \midrule
        DemoDICE~\cite{kim2021demodice} & 76.4 & 78.8 & 39.6 & 16.0 & 8.8  \\
        DWBC~\cite{xu2022discriminator} & 32.8 & 64.0 & 26.8 & 10.4 & 4.0  \\
        ISW-BC~\cite{li2024imitation} & 79.2 & 82.4 & 45.4 & 26.2 & 18.8 \\
        \midrule
        SSDF-base & 83.6 & 84.0 & 44.4 & 26.0 & 16.0 \\
        SSDF & \textbf{85.6} & \textbf{88.0} & \textbf{50.4} & \textbf{29.6} & \textbf{28.4} \\
        \bottomrule
        \end{tabular}
    }
\vspace{-10pt}
\end{table}

\textbf{Real-world Experiments.}
We follow Droid~\cite{khazatsky2024droid} to build our real-world robotic arm environment. 
As shown in Figure~\ref{fig:task_setting}, we use a Franka Panda 7-DoF robot arm.
The input RGB images $o \in \bbR^{2*270*480*3}$ are from  two external fixed-view Zed 2 stereo cameras.
We collect the dataset through human demonstration. 
For each task, we randomly place the objects within a specified area and collect 30 success trajectories as the expert dataset and 70 failed trajectories as the imperfect dataset. 
We record the RGB images from two camera views and robot proprioceptive states, e.g., joint position $m \in \bbR^{7}$. 
Our model predicts the 6D pose, including position $(x,y,z)$ and rotation $(roll, pitch, yaw)$.

\subsection{Evaluation Results}

\begin{table}[t]
    \centering
    \caption{Comparisons in terms of success rates of the five tasks on single-arm Franka robot.}
    \label{tab:franka_results}
    \resizebox{\columnwidth}{!}{
        \begin{tabular}{c|ccccc|c}
        \toprule
        \multirow{2}{*}{Method} & Place & Open & Open & Stack & Arrange & \multirow{2}{*}{Avg.} \\ 
        & Tennis & Drawer & Box & Block & Flower & \\
        \midrule
        BeT~\cite{shafiullah2022behavior} & 20 & 30 & 25 & 10 & 0 & 17 \\
        ACT~\cite{zhao2023learning} & 45 & 45 & \textbf{65} & 20 & 20 & 39 \\
        Diffusion Policy~\cite{cheng2023diffusion} & 35 & 35 & 60 & 25 & 20 & 35 \\
        ISW-BC~\cite{li2024imitation} & 40 & 45 & \textbf{65} & 20 & 15 & 37 \\
        \cmidrule(lr){1-7}
        SSDF+ACT & \textbf{50} & \textbf{55} & \textbf{65} & \textbf{30} & \textbf{25} & \textbf{45} \\
        \bottomrule
        \end{tabular}
    }
\vspace{-10pt}
\end{table}

\textbf{Results on Simulation.}
We comprehensively compared SSDF to many state-of-the-art offline imitation learning algorithms including 4 types. \textbf{1) IL methods that only use expert demonstrations}: Behavior Cloning (BC)~\cite{pomerleau1988alvinn}, Transformer-based Behavior Cloning (TF-BC)~\cite{vaswani2017attention}, Trajectory Transformer (TT)~\cite{janner2021offline}, and Behavior Transformer (BeT)~\cite{shafiullah2022behavior}. \textbf{2) Pretraining methods that pre-train model with mixed data and then finetune using expert data}: Perception-Action Causal Transformer (PACT)~\cite{bonatti2023pact}, Masked Trajectory Models (MTM)~\cite{wu2023masked}, and Self-supervised Multi-task pretrAining with contRol Transformer (SMART)~\cite{sun2023smart}.
\textbf{3) Goal-conditioned methods that take task completion signal as input}: Decision Transformer (DT)~\cite{chen2021decision}, and Conditional Behavior Transformers (C-BeT)~\cite{cui2022play}.
\textbf{4) IL methods that rank imperfect data:} DemoDICE~\cite{kim2021demodice}, Discriminator-Weighted Behavioral Cloning (DWBC)~\cite{xu2022discriminator}, and Importance Sampling Weighted Behavior Cloning (ISW-BC)~\cite{li2024imitation}.
To verify the effectiveness of the pre-training process, we trained \textbf{SSDF-base} that only leverages the imperfect demonstrations in the pre-training process. 
We trained all methods for 100k gradient steps and evaluated 50 episodes every 5k steps.
The final results are calculated as the average success rates of the last 5 checkpoints over 3 random seeds. 

Table~\ref{tab:maniskill-main-result1} shows that SSDF outperformed all baselines consistently on all 5 tasks by a large margin.
For IL methods that rank imperfect data~\cite{kim2021demodice,xu2022discriminator,li2024imitation}, once they assign high weights to low-quality data points, it in turn interferes with the results of the training. 
For example, DWBC only achieved a success rate of 32.8 on the PickCube task. This is due to the difficulty of the discriminator in predicting accurate weights for high-dimensional image inputs, which drastically reduced the success rate.
In contrast, SSDF can improve performance by accurately calculating the quality scores and filtering out high-quality demonstrations.

\textbf{Results on Single-arm Franka Robot.}
For the single-arm franka robotic tasks, we compared 4 state-of-the-art methods that are BeT~\cite{shafiullah2022behavior}, Action Chunking with Transformers (ACT)~\cite{zhao2023learning}, Diffusion Policy~\cite{cheng2023diffusion}, and ISW-BC~\cite{li2024imitation}.
We trained all methods for 200 epochs and evaluated the final checkpoints.
As shown in Table~\ref{tab:franka_results}, we observe that SSDF+ACT achieves average success rates of 45\% on these 5 tasks, which outperforms ACT by 6\%.
This increase in success rate shows that SSDF can select high-quality imperfect data, expand the original expert dataset, and thus improve the training of models.

\subsection{Ablation Study in Simulation}

\textbf{Impact of different self-supervised objectives.}
On the StackCube task, we evaluated SSDF with different combinations of the 3 self-supervised objectives in training step 1.
In Table~\ref{tab:self-sup-loss}, as we continue to increase the self-supervised objectives, SSDF can achieve higher success rates. 
This is because different self-supervised learning losses enable the model to learn versatile features, allowing it to compute quality scores for imperfect data accurately.

\begin{table}[t]
    \centering
    \caption{Ablation study on the different self-supervised objectives used in the training step 1 on the StackCube task.}
    \label{tab:self-sup-loss}
    \resizebox{0.95\columnwidth}{!}{
        \begin{tabular}{c|cccccc|c}
        \toprule
        StackCube & 1 & 2 & 3 & 4 & 5 & 6 & SSDF \\ 
        \midrule
        MTP & \ding{51} &  & & \ding{51} & \ding{51} &  & \ding{51}  \\
        TR &  & \ding{51} & & \ding{51} &  & \ding{51} & \ding{51} \\
        AA &  &  & \ding{51} &  & \ding{51} & \ding{51} & \ding{51} \\
        \midrule
         Result & 75.4 & 72.0 & 78.8 & 80.6 & 84.2 & 83.8 & \textbf{88.0} \\
        \bottomrule
        \end{tabular}
    }
\vspace{-5pt}
\end{table}
 
\begin{table}[t]
    \centering
    \caption{Comparisons of different quality score threshold $\beta$ on the StackCube task.}
    \label{tab:maniskill-beta-result}
    \begin{tabular}{l|ccccccc}
    \toprule
    Threshold $\beta$ & 0.0 & 0.50 & 0.70 & 0.80 & 0.90 & 0.95 & 0.99 \\ 
    \midrule
    success rate & 17.2 & 21.6 & 43.8 & 66.4 & 88.0 & 84.4 & 83.6 \\
    \midrule
    reserved data & 180k & 57k & 35k & 16k & 8.8k & 3.2k & 0.3k \\
    \bottomrule
    \end{tabular}
\vspace{-6pt}
\end{table}

\begin{table}[t]
    \centering
    \caption{Comparisons of different similarity functions on the StackCube task.}
    \label{tab:maniskill-func-result}
    \begin{tabular}{l|cccc}
    \toprule
     & Seg.+$Cosine$ & Last+$Cosine$ & Seg.+$L_2$ & Last+$L_2$ \\ 
    \midrule
    success rate & 77.2 & 83.2 & 82.0 & 88.0 \\
    \bottomrule
    \end{tabular}
\end{table}

\textbf{Impact of quality score threshold $\beta$.}
As shown in Table~\ref{tab:maniskill-beta-result}, on the StackCube task, we can observe that as the quality score threshold value $\beta$ increases, the reserved demonstrations become less and less.
The highest success rate on the StackCube task is with $\beta$ as 0.90.
When $\beta$ is less than 0.90, the performance increases as $\beta$ increases because more and more low-quality demonstrations are discarded. 
When $\beta$ is larger than 0.90, we argue that the performance drops as $\beta$ increases because more high-quality demonstrations are discarded, degrading the task performance.


\begin{figure}[tp]
\centering
\includegraphics[width=0.95\linewidth]{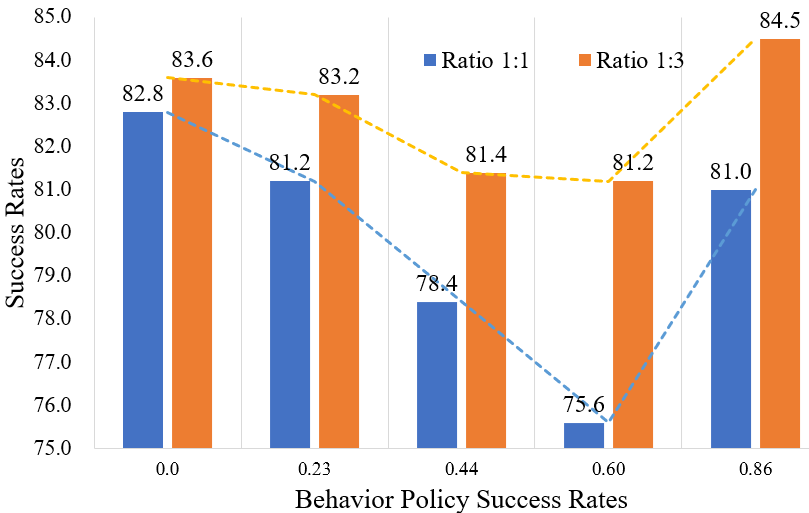}
\caption{Comparisons of different quantity and quality of the imperfect data. The x-axis represents the success rates of the behavior policies that collect the imperfect data. The y-axis shows the success rates of SSDF. The green and red lines represent the ratios of the expert and imperfect data, which are 1:1 and 1:3, respectively.}
\label{fig:stack_abl_qq}
\vspace{-4pt}
\end{figure}

\textbf{Impact of different similarity calculation functions.}
We examined $L_2$ distance and $Cosine$ similarity distance with the whole segment features denoted as ``Seg." and the last time step feature denoted as ``Last" on the StackCube task.
As shown in Table~\ref{tab:maniskill-func-result}, ``Last+$L_2$" achieved the highest performance, while the results of using cosine similarity are relatively lower.
%
We argue that the last time step of the trajectory segment is the most important step for predicting the next action, and the attention mechanism in the transformer has extracted the historical information.

\textbf{Impact of the imperfect demonstration quantity and quality.}
For the StackCube task, we collected five different level imperfect demonstrations using different behavior policies with success rates of 0.00, 0.23, 0.44, 0.60, and 0.86, respectively .
For each level, we collected imperfect demonstrations in two ratios, 1:1 and 1:3, where the former 1 represents the 300 expert demonstrations.

Figure~\ref{fig:stack_abl_qq} shows the success rates with different imperfect demonstration quantities and qualities on the StackCube task.
An interesting finding is that those low-quality and high-quality imperfect demonstrations can benefit the performance compared to the medium-quality imperfect demonstrations.
We believe that low-quality demonstrations contain rich explorations of the environment and therefore can help extract more informative features in the pre-training process.
Then they are discarded in the fine-tuning stage and do not interfere with the results.
In contrast, high-quality imperfect demonstrations directly help the process of the fine-tuning stage. 
The medium-quality demonstrations neither provide rich environmental information nor directly enhance policy learning, so they achieve the most minor improvement.




\section{Conclusion}
This paper explores how to leverage imperfect demonstration to improve policy learning without reward information and online exploration.
To this end, we propose a novel \textbf{S}elf-\textbf{S}upervised \textbf{D}ata \textbf{F}iltering framework (SSDF) that calculates accurate quality scores using a pre-trained transformer and then does weighted behavior cloning with the high-quality imperfect demonstrations.
The extensive experimental results in simulation and real-world applications demonstrated that SSDF can accurately select high-quality imperfect demonstrations to boost the final performance.

\clearpage
\bibliographystyle{IEEEtran}
\bibliography{reference}

\end{document}